\documentclass[runningheads,a4paper]{llncs}

\let\subparagraph\paragraph
\usepackage{titlesec}
\usepackage{amssymb}
\usepackage{graphicx}
\usepackage{url}
\usepackage{bibnames}
\usepackage{wrapfig}
\usepackage{url}
\urldef{\mailsa}\path|{rw37, ms5267, ajj37}@drexel.edu|
\usepackage{subcaption}

\urldef{\mailsc}\path| |    
\newcommand{\keywords}[1]{\par\addvspace\baselineskip
\noindent\keywordname\enspace\ignorespaces#1}

\titlespacing\section{0pt}{18pt plus 8pt minus 2pt}{4pt plus 2pt minus 2pt}
\titlespacing\subsection{0pt}{12pt plus 6pt minus 4pt}{6pt plus 4pt minus 2pt}
\titlespacing\subsubsection{0pt}{10pt plus 4pt minus 2pt}{4pt plus 2pt minus 2pt}

\begin{document}

\mainmatter  

\title{Knowledge-based XAI through CBR: There is more to explanations than models can tell}

\author{Rosina O. Weber\textsuperscript{\rm 1},
        Manil Shrestha\textsuperscript{\rm 2}, and
        Adam J Johs\textsuperscript{\rm 1}
}
\authorrunning{Weber et al.}
\titlerunning{Knowledge-based XAI through CBR}
\institute{Information Science\textsuperscript{\rm 1}, Computer Science\textsuperscript{\rm 2}\\ Drexel University, Philadelphia, PA 19104\\
{\{rw37, ms5267, ajj37\}}@drexel.edu
}

\maketitle
\begin{abstract}
The underlying hypothesis of knowledge-based explainable artificial intelligence is: the data required for data-centric artificial intelligence agents (\textit{e.g.,} neural networks) are less diverse in contents than the data required to explain the decisions of such agents to humans. The idea is that a classifier can attain high accuracy using data that express a phenomenon from one perspective whereas the audience of explanations can entail multiple stakeholders and span diverse perspectives. We hence propose to use domain knowledge to complement the data used by agents. We formulate knowledge-based explainable artificial intelligence as a supervised data classification problem aligned with the CBR methodology. In this formulation, the inputs are case problems composed of both the inputs and outputs of the data-centric agent, and case solutions, the outputs, are explanation categories obtained from domain knowledge and subject matter experts. This formulation does not typically lead to an accurate classification, preventing the selection of the correct explanation category. Knowledge-based explainable artificial intelligence extends the data in this formulation by adding features aligned with domain knowledge that can increase accuracy when selecting explanation categories.

\keywords{explainable artificial intelligence, knowledge, expertise, interpretable machine learning, case-based reasoning}

\end{abstract}

\setlength{\floatsep}{3.0pt} 
\setlength{\textfloatsep}{16.0pt plus 2.0pt minus 4.0pt} 
\setlength{\intextsep}{2.0pt} 
\setlength{\abovecaptionskip}{2.0pt} 
\setlength{\belowcaptionskip}{2.0pt} 

\begin{figure}[h]
\centering
\includegraphics[width=12.4cm,keepaspectratio]{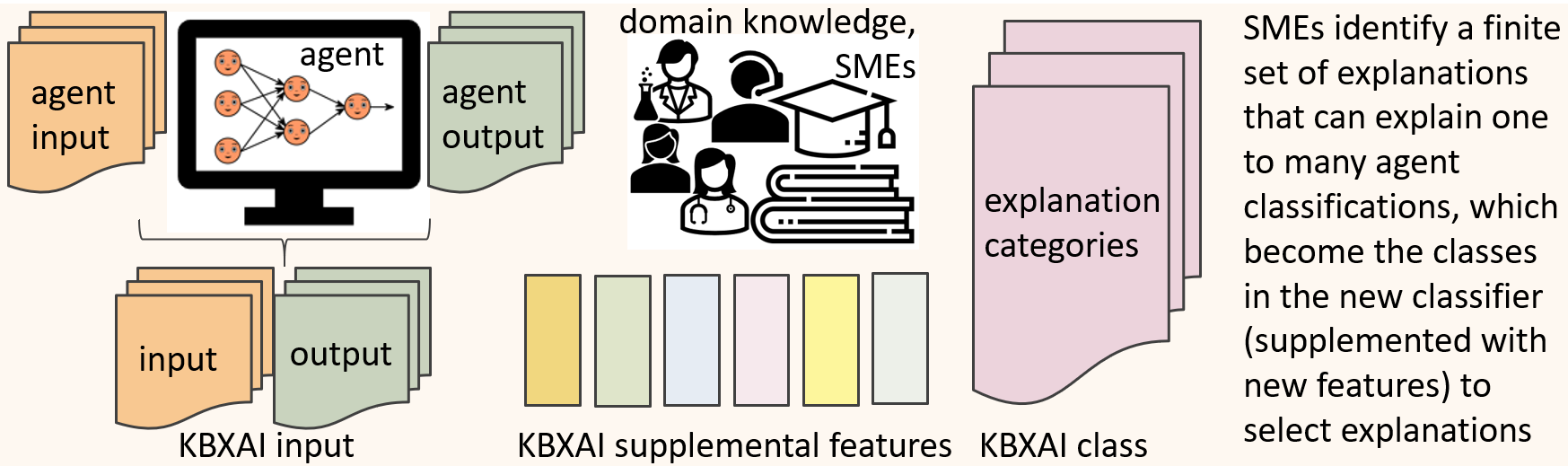} 
\caption{Overview of KBXAI: Step 1 obtains explanation categories; Step 2 supplements the data with new features to select explanations to explain classifications}
\label{overview}
\end{figure}

\section{Introduction}

\noindent
Explainable artificial intelligence (XAI) (\textit{e.g}., \cite{adadi:2018}) is a sub-field of artificial intelligence (AI) research that arose with substantial influence from interpretable machine learning (IML) (\textit{e.g.,} \cite{doshivelez2017rigorous,khanna2019interpreting}). The focus of IML has always been (\textit{e.g.,} \cite{andrews1995survey}) to promote interpretability to ML experts who need to comprehend how concepts are learned to advance the state of the art. The research in IML can be broadly categorized as \textit{feature attribution} (\textit{e.g.,} \cite{simonyan2013deep}), \textit{instance attribution} (\textit{e.g.,} \cite{shrikumar2017learning}), and \textit{example-based} (\textit{e.g.,} \cite{nugent2005case}). XAI has been proposed \cite{Gunning_Aha_2019} with a focus on explainability to users---despite this distinction, most methods for XAI are limited to considering only the interpretability of models (\textit{e.g.,} \cite{simonyan2013deep,shrikumar2017learning,folke2021explainable}).

The literature in XAI implies that seeking explanations solely within AI models is insufficient. This implication is supported by various authors (\textit{e.g.,} \cite{lim2012improving,nunes2017systematic} who have provided contents of explanations not available in models (See Section \ref{extype} for detailed discussion). Knowledge-based explainable artificial intelligence (KBXAI) is an approach to XAI that seeks to bridge this gap by acquiring knowledge for explanations from domain knowledge and subject matter experts (SMEs). In this paper, we introduce and describe how to implement KBXAI with case-based reasoning (CBR). KBXAI (See Fig. \ref{overview}) is implemented in two steps: 1) defining explanation categories, and 2) case extension learning.

The next section presents background and related works. With the goal of examining challenges and opportunities brought to bear with the introduction of KBXAI, we illustrate and discuss KBXAI in three problem contexts, each with different data types---tabular, image, and text.

\section{Related Works}

\subsection{Explanation Types}\label{extype}
\noindent
In this section, we review works where authors proposed various explanation contents for use in explanations of intelligent agents. This review is not exhaustive but illustrates how the breadth of explanation contents extends beyond models and data.

Lim \cite{lim2012improving} describes a taxonomy of \emph{explanation types} that include \emph{situation, inputs, outputs, why, why not, how, what if, what else, visualization, certainty}, and \emph{control}. The item \emph{input} refers to external sources an intelligent agent may have used. In the credit industry, for instance, companies purchase hundreds of thousands of credit profiles of unidentified applicants that are not directly considered in the explanations. Another item is \emph{situation}, which Lim (ibid.) exemplifies with an industrial process where an anomaly is presented, triggering an agent's decision. Lim (ibid.) states that some users would like explanations to include what the \emph{normal} process was prior to the anomalous event.

Nunes and Jannach \cite{nunes2017systematic} conducted a systematic review of the literature toward understanding the characteristics of explanation content provided to users across multiple intelligent systems. Explanations proposed in the literature were qualitatively coded to identify the types of contents communicated in explanations — 17 types of explanation contents were identified and grouped as: 1) user preferences and inputs, 2) decision inference process, 3) background and complementary information, and 4) alternatives and their features. Nunes and Jannach (ibid.) consider multiple forms of background and complementary information, mostly external to the data or knowledge used by intelligent agents---\textit{e.g.,} the background information a human would necessitate for a classification instance; this dovetails with Lim's \emph{inputs} and \emph{situation} explanation types.

Chari et al. \cite{chari2020explanation} propose a taxonomy of approaches and algorithms to support user-centered AI system design. This taxonomy includes two components of scientific explanations divorced from AI models and the data used by AI agents: the scientific method and evidence from the literature. Additional contributions to explanation types are found in \cite{gedikli2014should,vilone2020explainable}.

\subsection{Three main categories of IML and XAI methods}
\noindent
The three main categories of IML and XAI methods are feature attribution \cite{simonyan2013deep,shrikumar2017learning,lundberg2017unified,bach2015pixel,ribeiro2016should,sundararajan2017axiomatic}, instance attribution \cite{khanna2019interpreting,koh2017understanding,yeh2018representer,mercier2020interpreting,barshan2020relatif}, and example-based \cite{nugent2005case,folke2021explainable,kim2014bayesian,kenny2019twin,kenny2021explaining}---all predicated on obtaining explanations from an agent's model. Attribution methods explain model behavior by associating an input solved by an agent to elements of the model used by that agent, either by looking at the instance features (\textit{i.e.,} feature attribution) or by looking at each instance as an integral component (\textit{i.e.,} instance attribution). KBXAI neither employs attribution nor prescribes reliance on examples. KBXAI may use features from the model, but knowledge external to the model is required, signifying better alignment with an additional category of model-extrinsic methods. In the XAI categorization as \textit{intrinsic} and \textit{post-hoc}, KBXAI is \textit{post-hoc} because it is implemented after rather than contemporaneously to the agent as with intrinsic methods (\textit{e.g.,} \cite{li2018deep}).

Feature attribution methods are relatively easy to compute and have risen in popularity (\textit{e.g.,} \cite{simonyan2013deep,bach2015pixel,ribeiro2016should,lundberg2017unified,shrikumar2017learning,sundararajan2017axiomatic}). Among such methods are LIME \cite{ribeiro2016should}, which creates perturbations and then fits them to a linear regression to explain a point that participates in the straight line with its coefficients. Saliency methods \cite{simonyan2013deep,shrikumar2017learning,bach2015pixel} are widely used to explain image models because such methods afford the construction of heat maps that emphasize regions (\textit{i.e.,} features) of an image where weights are higher. Another prevailing method is SHAP \cite{lundberg2017unified}, which adds rigor from Shapley values to feature attribution based on perturbations. Such methods have been criticized for producing the same explanation despite noise added to the data or changes made to the models \cite{adebayo2018sanity,kindermans2017unreliability}. Feature attribution have been found to not work in neural architectures that use a memory \cite{koul2018learning}.

Instance attribution methods provide the instances associated with a decision \cite{khanna2019interpreting,koh2017understanding,yeh2018representer,mercier2020interpreting,barshan2020relatif}. These methods have been shown to have multiple uses such as debugging models, detecting data set errors, and creating visually indistinguishable adversarial training examples \cite{koh2017understanding}. In addition to being computationally expensive \cite{khanna2019interpreting}, there are other criticisms to these methods--\textit{e.g}., attributed instances are often outliers and the sets of instances attributed to different samples have substantial overlap \cite{barshan2020relatif}. 
\label{marker}
Methods that select training instances based on some similarity concept as the basis for explanations are known as example- or prototype-based (\textit{e.g.,} \cite{nugent2005case,folke2021explainable,kim2014bayesian,kenny2019twin,kenny2021explaining}). Example-based methods are relatively easy to compute and have been successful in user studies \cite{nugent2005case,kenny2021explaining,folke2021explainable}; the core problem with such methods is the absence of attribution.

\subsection{Domain knowledge in XAI}
\noindent
Domain knowledge has been used as part of explanation for recommender systems (\textit{e.g.,} \cite{zanker2010knowledgeable}), expert systems (\textit{e.g.,} \cite{swartout1993explanation,wick1989reconstructive}), and CBR systems (\textit{e.g.,} \cite{bergmann1993explanation,aamodt1993explanation,weber2018}. For scientific insights and scientific discoveries, domain knowledge is considered i) a prerequisite for attaining scientific outcomes, ii) pertinent to enhancing scientific consistency, and iii) necessary for explainability \cite{roscher2020explainable}. In the biomedical domain, Pesquita \cite{pesquita2021towards} proposed augmenting post-hoc explanations with domain-specific knowledge graphs to produce \emph{semantic explanations}.

Contextual decomposition explanation penalization \cite{rieger2020interpretations} permits insertion of domain knowledge into deep learning with the aim of mitigating false associations, rectifying errors, and generalizing to other methods of interpretability; examples of incorporable domain knowledge range from human labeled ground truth explanations for every data point, to the importance of various feature interactions. \cite{schramowski2020making}'s explanatory interactive learning method leverages human-in-the-loop revision to align model explanations with the knowledge of the expert in-the-loop.

\begin{figure} [t]
\centering
\includegraphics [width=12.4cm,keepaspectratio] {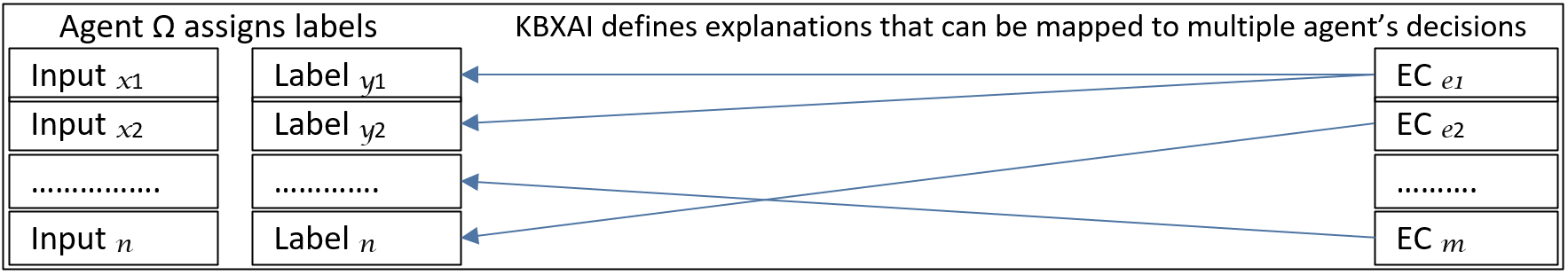}
\caption{Explanations are categorized and mapped to agent's input-output pairs}
\label{defineEC}
\end{figure}

\section{Knowledge-based explainable AI}
\noindent
Based on the premise that data used by an agent is to be supplemented, we formulate KBXAI as a problem where input data is given to an agent to execute an intelligent task (for simplicity, henceforth this task is referred to as {\it classification}). Consider a classifier agent $\Omega$ using training instances from an input space $X$ to an output space $Y$ that uses labeled training instances $z_1,...,z_n \in Z$ where $z_i = (x_i,y_i)  \in  X \times Y$. 

As introduced in \cite{weber2018}, KBXAI has two main steps: 1) defining explanation categories, and 2) case extension learning. Fig. \ref{defineEC} shows the first step when KBXAI uses domain knowledge to define a finite set of explanation categories (EC) $e \in EC$, which are defined by a mapping def: $Z \times EC \rightarrow$ 0 or 1. We refer to these explanations as categories because they are meant to explain one to many classifications. Within KBXAI, the explanations are textual even when explaining images to facilitate incorporation of supplemental features.

This step creates a new classification problem, which we formulate as case-based. The case problems are the inputs and outputs (i.e., or labels) of the agent. The case solutions are the explanation categories that we wish to select for each agent's classification. This formulation produces low accuracy because it is typically indeterminate. The reason for this is that the explanations include contents that are not in the data and model used by the agent $\Omega$. The goal of KBXAI is \emph{to successfully select the correct explanation category for a given input-output pair}. This prompts the need for the next step.

\begin{figure}[t]
\centering
\includegraphics [width=12.4cm,keepaspectratio] {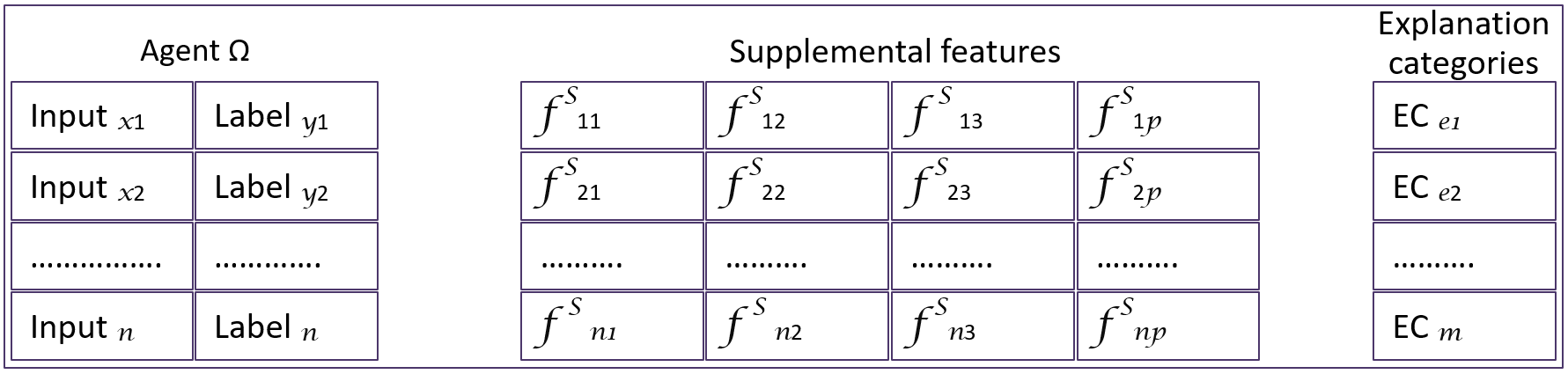} 
\caption{Supplemental features contribute to select accurate explanation categories}
\label{supp}
\end{figure}

The second step, case extension learning, is when KBXAI supplements the data by proposing and evaluating supplemental features. The aim is to find features that improve the baseline accuracy to successfully select the correct explanation category for a given input-output pair (see Fig. \ref{supp}). Proposing features from domain knowledge represents a knowledge engineering step.

\subsection{Case-based implementation}
\noindent
We implement case extension learning with CBR. There are two main reasons for not adopting a data-centric approach like neural networks (NN), namely, lack of transparency and sample distribution discrepancy. When comparing the performance of an NN with and without one or more features, if the testing data used are the same for testing both variations then it places the testing and training at different distributions. This does not conform with the machine learning (ML) principle that testing and training must come from the same distribution (\textit{e.g.,} \cite{hooker2018benchmark,dabkowski2017real,fong2017interpretable}). CBR is transparent and allows evaluation of features without violating the ML principle. Through ablation using weighted k-Nearest Neighbor (kNN) and leave-one-out cross validation (LOOCV), when a feature is included, all instances that are left out in LOOCV include such feature, when excluded, the instances do not include it.

Implementing case extension learning with CBR through ablation is as follows. With the problem formulation as depicted in Fig. \ref{defineEC}, we used ReliefF \cite{kononenko1997overcoming} to learn weights with local similarity as either a binary (\textit{i.e.,} equal vs unequal) function or, when applicable, a function that computes the difference between values and normalizes based on the range of observed values. Average accuracy is computed with LOOCV. Baseline accuracy is computed with the agent's inputs and outputs, and explanation categories; this is before adding any supplemental features. We do not always we have access to the representations of the input to the agent. In these situations, we consider input as a nominal feature. Case extension entails proposing candidate supplemental features and evaluating how they impact overall average accuracy with respect to the baseline accuracy. The supplemental features are evaluated one at a time and then in aggregation. Supplemental features may not be independent to the features that come from the agent's input, which is fine because we analyze their impact and keep the ones that better contribute to increasing accuracy. When features are redundant, they do not increase accuracy proportionally when combined.

Next we describe studies applying KBXAI in three data sets using different data types. The goal of these studies is to assess potential challenges for KBXAI. Each data set was obtained differently and the knowledge used to complement the problems also varies. None of the studies represent a complete real-world scenario where KBXAI could be fully deployed. The increments in accuracy shown are modest. Considering that we found supplemental features that caused accuracy to increase is what demonstrates the KBXAI hypothesis has potential.

\subsection{KBXAI in Tabular data}
\noindent
This synthetic data set is a binary classification with labels accept and reject \cite{amiri2020data}. This data has 54 instances and three features with four allowable values each. The first feature, job stability (X1), corresponds to the job status of the applicant. This feature has integer values [2, 5], where 2 means lack of a job, and values 3, 4, and 5, respectively, that applicant has a job for less than one year, less than 3 years, or more than 3 years. The second feature is credit score, credit score (X2), and has values [0, 3], meaning less than 580, 650, 750 and more than 750. The third feature is the ratio of debt payments to monthly income (X3), with values [0, 3], meaning less than 25\%, 50\%, 75\% and more than 75\%.

The agent is a NN architecture with four hidden layers and 512 neuron and ReLU activation layers, ending with a sigmoid activation layer. The loss function is binary cross-entropy and the optimizer used is gradient descent. The classifier reached 100\% accuracy, which is likely overfit given that we did not separate data because of the small number of samples.

\begin{figure}[h]
\centering
\includegraphics[width=12.4cm,keepaspectratio]{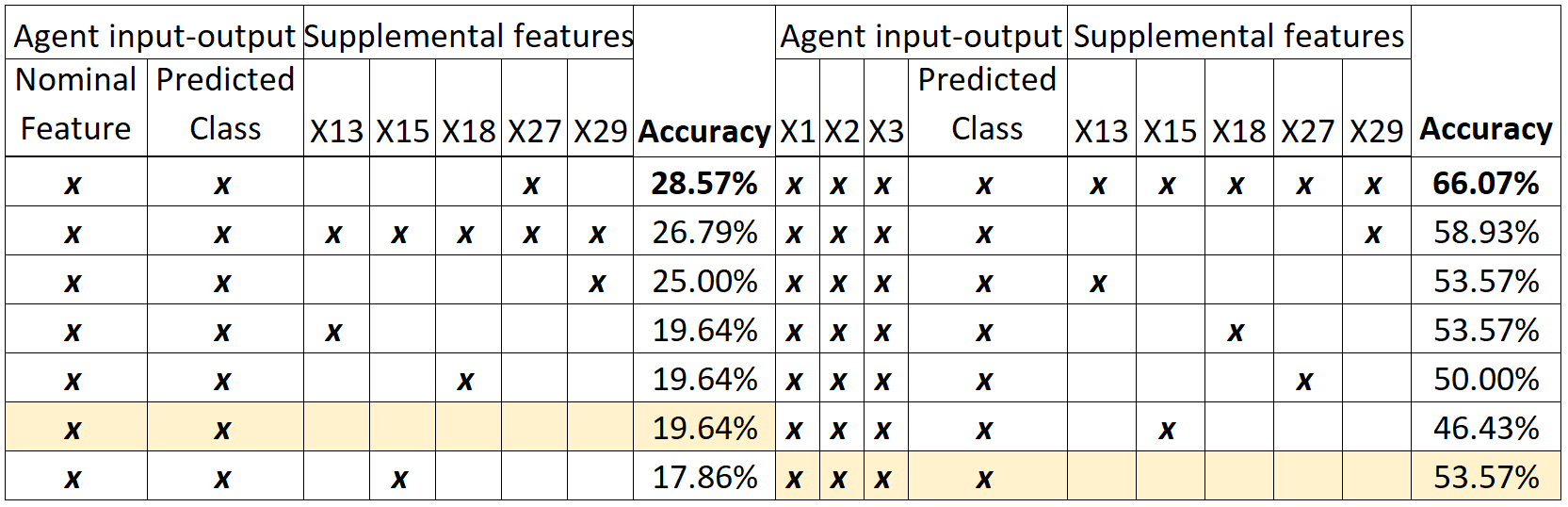} 
\caption{Case extension learning for tabular data with nominal input features vs. agent's input. Baseline accuracy is indicated in yellow and maximum in bold}
\label{tabular}
\end{figure}

To identify explanation categories, these authors used their own knowledge of credit assessment combined with online resources to identify 15 explanation categories that align to the 54 instances. The explanation categories are hypothetical rules combining feature values in both accept and reject classes. Some example explanation categories are, "\emph{with lowest credit score, either job condition and debt have to be excellent or both very good for acceptance"}; "\emph{no job and credit score is not excellent then reject"}; and "\emph{despite no job, credit score is excellent then accept"}.

Fig. \ref{tabular} summarizes two examples of case extension learning with this data set. On the left of Fig. \ref{tabular}, we implement the agent's input as a nominal feature, on the right we use the three features used by the agent. The baseline accuracy is 17.8\% and 53.5\%, respectively. This is not surprising, given the agent's input are the basis of the agent's learning.  

We created 29 features by combining values of subset of features and decisions. We only describe those that improved accuracy. Feature X3 is obtained with a function that assigns 1 when despite the debt-income ratio being greater than 75\%, the applicant  still is approved for credit. For Feature X15, the function assigns 1 if debt-income ratio is less than 25\% and the result is approved. Feature X18 is the same as X13 for rejected applications. Feature X27 is valued 1 when credit score is less than 650 and the decision is accept. Feature X29 requires credit score to be below 750 and debt-income ratio not to be $<$25\% when the class is reject to receive value 1 and is zero otherwise. 

\begin{figure}[h]
  \begin{subfigure}{0.49\textwidth}    \includegraphics[width=\linewidth]{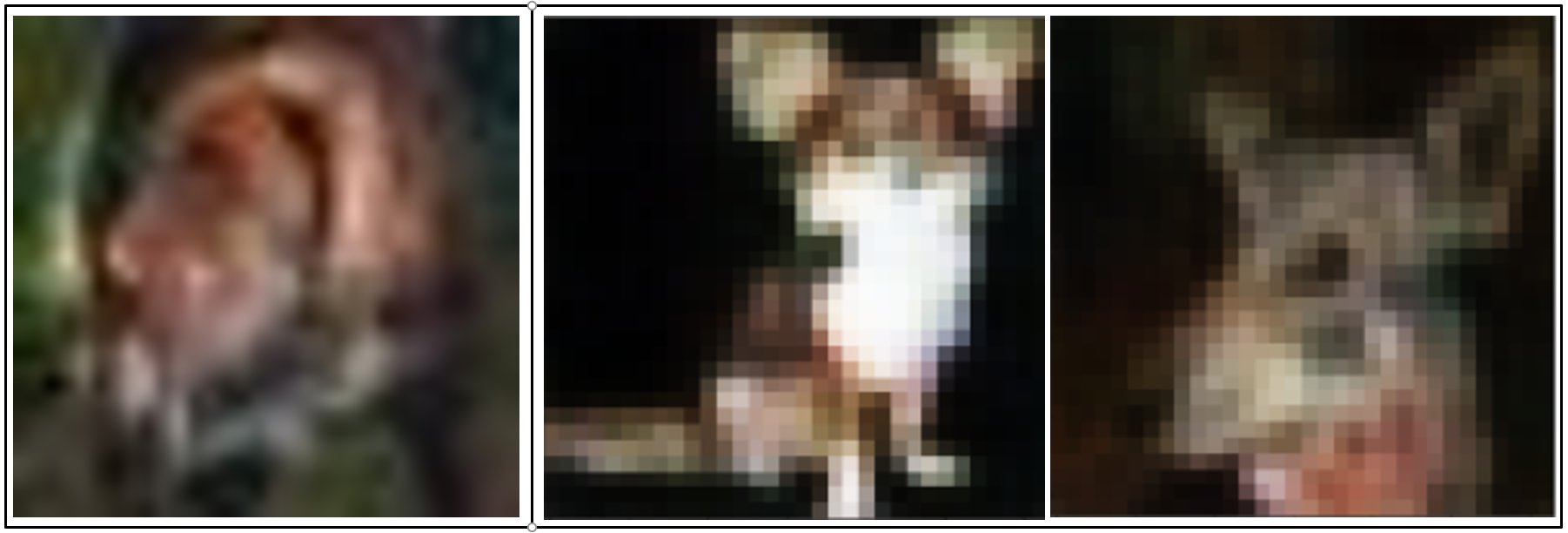}
    \caption{ } \label{fig:dogs1}
  \end{subfigure}%
  \hspace*{\fill}   
  \begin{subfigure}{0.49\textwidth}
    \includegraphics[width=\linewidth]{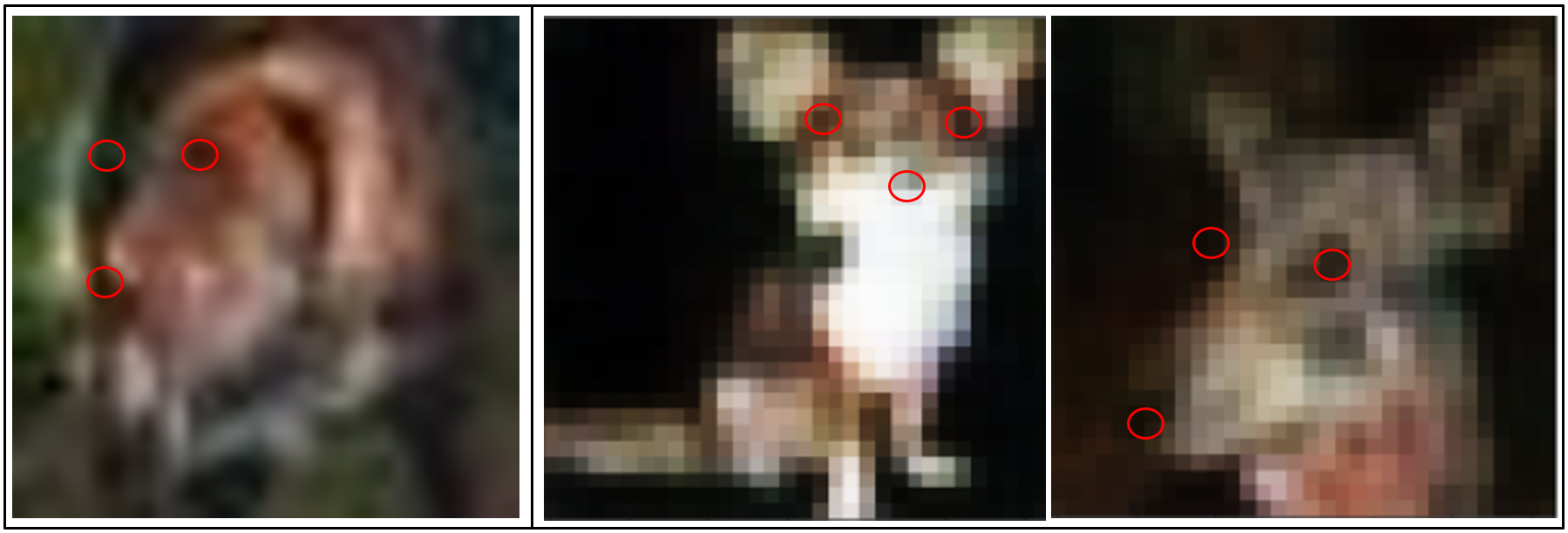}
    \caption{} \label{fig:dogs2}
  \end{subfigure}%

\caption{\textbf{(a)} Example explanation category: On the left is input image with true label dog; on the right, two images with high median similarity, originally misclassified by the agent that are selected as the explanation category for the input
\textbf{(b)}The feature Triangular Markings is valued at 1 when authors agreed they saw the three dots and zero otherwise
} \label{dogs}
\end{figure}

With the same supplemental features, the accuracy improves about 60\% when using nominal values for input, and about 23\% when the input features are used (Fig. \ref{tabular}). We also note the combinations of supplemental features reveal different performances in these two executions, potentially suggesting that some of the supplemental features are redundant with respect to the agent's input. Note how they improved accuracy when considered alone and in combination with other features. The best performing feature changes from X27 to X29 in the two executions.

\subsection{KBXAI in Image data} 
\noindent
The data for the study with images is a subset of CIFAR-10 \cite{Krizhevsky}. Out of 10 classes, we selected four, namely, dogs, trucks, cats, and horses. We formulated these data as a binary classification of dog or not dog. The entire data set has 5,000 images per class for training and 1,000 for testing. We trained a VGG-16 architecture that reached 85\% accuracy for the binary classification.

To identify explanation categories, we adopted an example-based strategy for selecting example images for explanations (See Section \ref{extype}). The strategy is to select images that are like the image whose classification we want to explain. The candidate images to be used for explanations are the false negatives produced by the binary VGG-16 architecture classifier. The false negatives are all images of the class dog that were misclassified (outliers). See example in Fig. \ref{fig:dogs1}.

\begin{figure}[t]
\centering
\includegraphics[width=0.67\textwidth,keepaspectratio]{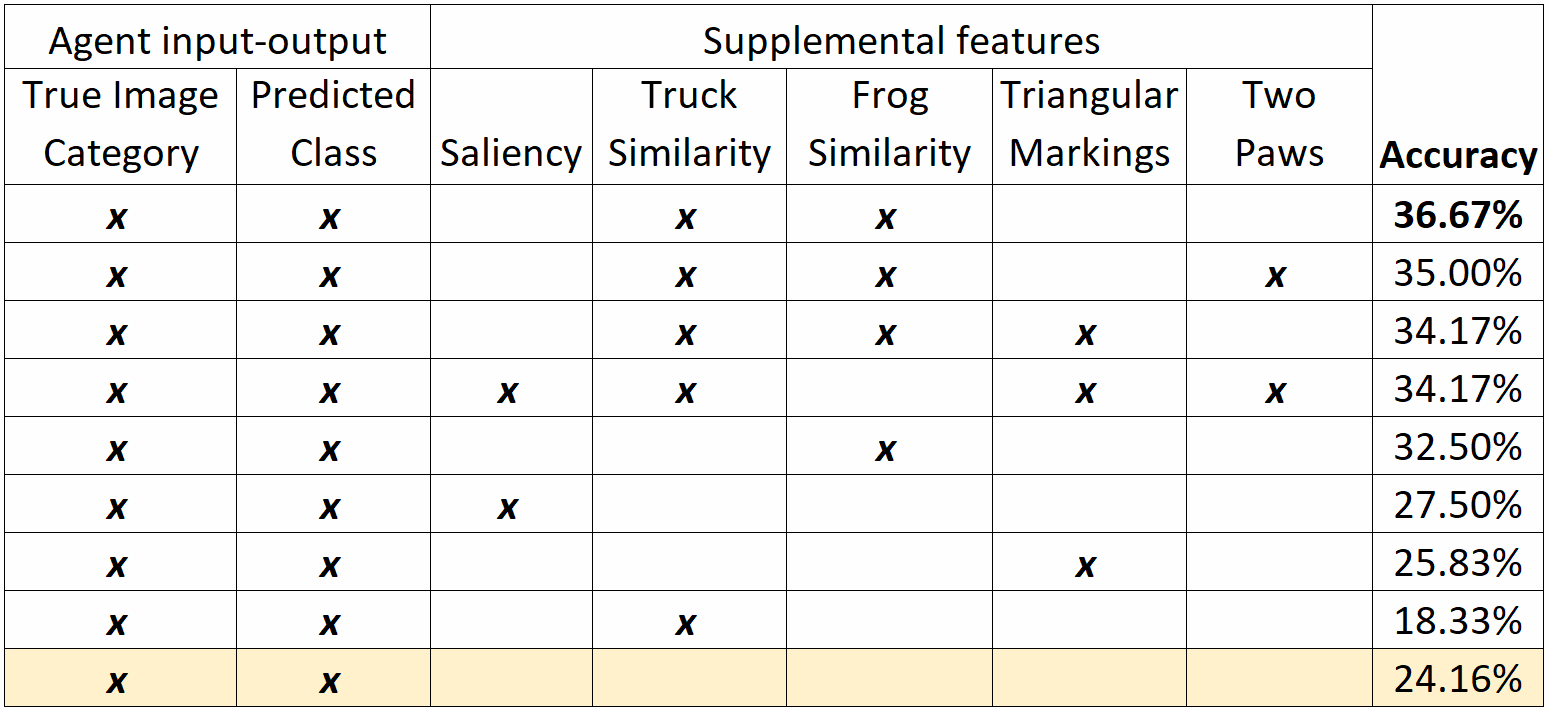} 
\caption{For image data, accuracy increased from 24\% to 36\%}
\label{image}
\end{figure}

To create explanation categories, we selected a subset of Cifar-10 test instances from the selected classes. For each instance, we computed the cosine between the embedding vectors of each test instance and all false negative dog images from the initial test excluding the image itself. To create the explanation category, we used two candidate images with the highest similarity score. Embedding vectors of images were created with an autoencoder.

After explanation categories are created, we now create the data set that maps the agent's input instances and their classification to their explanation category. Note this step was done as a proxy to having a subject matter expert selecting explanation categories. We used this approach because we did not want to have any of the authors interfere with this selection because we would select supplemental features. This mapping refers to identifying the correct explanation category for each instance. We selected explanation categories for each instance by computing the median value of the cosine similarity between the embedding vectors of the instances and the two images in the explanation category.
Once this step was completed, we removed duplicates from the resulting explanation categories. We then removed the explanation categories obtained with lower cosine values. We then examined the mapping of testing instances to explanation categories to select the explanation categories that explained more instances as a measure of their popularity. Finally, we took the 12 most popular explanation categories and randomly selected 10 testing instances mapped to each. The final data set has 120 instances and 12 explanation categories.

\begin{wrapfigure}{o}{0.41\textwidth}
\centering
\includegraphics[width=0.4\textwidth,keepaspectratio]{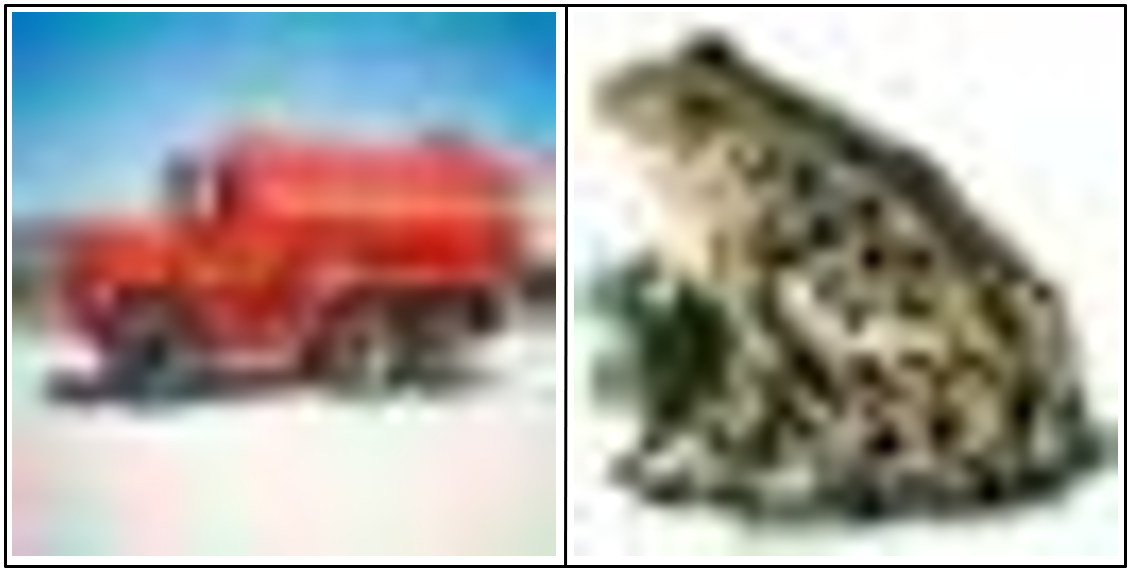} 
\caption{Images of a typical truck and a typical frog selected for features truck and frog similarity}
\label{typical}
\end{wrapfigure}

For images, we utilized both model and domain knowledge to propose features (Fig. \ref{image}). From the model, we computed the image’s saliency \cite{simonyan2013deep}. Saliency brought the 24\% baseline to 25\%. Truck similarity is a feature that assigns 1 to truck images and the cosine similarity between the embedding of image instance to embedding of a truck we selected as typical (See left of Fig. \ref{typical}). This feature is from the data as trucks are also part of the data set.

All other features are from commonsense knowledge, which here replaces domain knowledge. Frog similarity is the cosine similarity between embeddings of image instance and frog image we selected as typical of a frog (See left of Fig. \ref{typical}). Frog similarity alone increased accuracy from 24 to 32.5\%. The two last features were selected based on what the authors perceived in the pictures as possibly explaining the recognition of a dog. Note these images are so blurred that human accuracy is estimated to be around 90\% \cite{ho2018cifar10}. We found that dogs commonly have their eyes and nose as a triangle, marked in red in Fig. \ref{fig:dogs2}. Analogously, two paws refer to the images where the two front paws of the dog are clearly distinguishable.

\begin{wrapfigure}{o}{0.41\textwidth}
\centering
\includegraphics[width=0.4\textwidth,keepaspectratio]{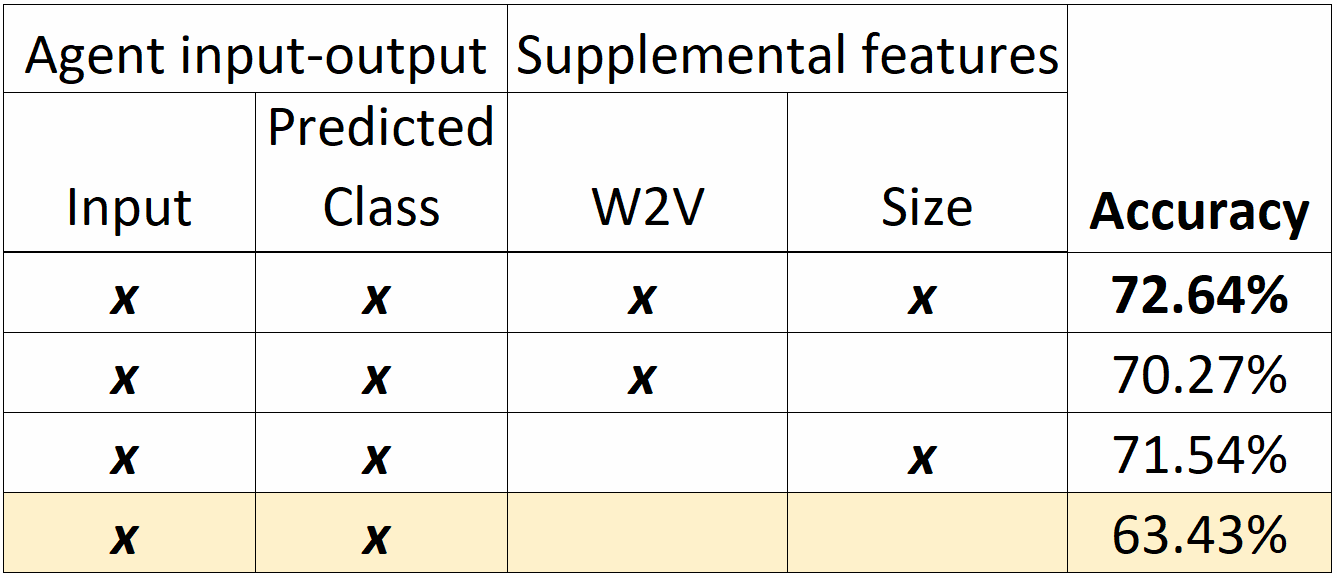} 
\caption{Starting from a high baseline of 63.43\%, two features increased accuracy to 72.64\%}
\label{textual}
\end{wrapfigure}

This example shows that the combination of features from the model and from commonsense knowledge together work better to increase accuracy. Individually, frog similarity showed the best performance. Note that frog images were not included in the data, making this a feature that is extrinsic to the model and the data.

\subsection{KBXAI in Textual data} 
\noindent
The KBXAI implementation with textual data was was previously published in \cite{weber2018}. The data set was built from a selection of 10 scientific articles. The agent used is a citation recommender \cite{bhagavatula2018content} that produces articles to be cited in the input article. We submitted the 10 articles as inputs to the recommender 10 times to create 100 cases for KBXAI. 
The explanation categories were learned from the domain of citation analysis. There are only two explanation categories, namely, \emph{background} and \emph{paraphrasing}.
Fig. \ref{textual} show results of case extension learning. The baseline of 63.43\% was increased to an accuracy of 72.64\% with only two features.
The baseline accuracy is higher than with previous data probably due to a binary selection of explanation categories. The improvement was only 14.5\%. Like in image data, only nominal features were used, which limits accuracy.

\section{Concluding remarks}
\noindent
Incorporation of knowledge engineering inevitably carries its problems such as difficulty to scale. However, KBXAI only requires knowledge acquisition to capture the additional perspectives that account to multiple stakeholders; it is not meant to acquire knowledge for an entire agent. The explanations tend to repeat and this is why we group them in categories. An open question is approaches where domain knowledge can be learned to be incorporated into KBXAI. The features we added to the tabular example are functions based on rules. This is aligned with explanation-based learning \cite{dejong1986explanation}, pointing to a future direction. 

The studies in this paper suggest it is necessary to leverage the representation used by the agent and not nominal features to represent the agent's input. This imposes on KBXAI, and consequently on CBR, that it should handle any type of representation. To overcome this, one direction is to adopt ANN-CBR twins \cite{kenny2019twin} and utilize the original representations. The next steps also include implementation in real-world scenarios and evaluation with humans. 

\subsubsection*{Acknowledgments} 
The authors thank the anonymous reviewers for suggestions to improve this paper. Support for the preparation of this paper was provided by NCATS, through the Biomedical Data Translator program (NIH award 3OT2TR003448-01S1). Authors Weber and Shrestha are also partially funded by DARPA-PA-20-02-06-POCUS-AI-FP-023.

\bibliographystyle{unsrt}
\bibliography{kbxai.bib}


\end{document}